\documentclass[10pt,twocolumn,letterpaper]{article}

\usepackage{cvpr}
\usepackage{times}
\usepackage{epsfig}
\usepackage{graphicx}
\usepackage{amsmath}
\usepackage{amssymb}
\usepackage{multirow}
\usepackage{float}
\usepackage{caption} 

\usepackage[pagebackref=true,breaklinks=true,letterpaper=true,colorlinks,bookmarks=false]{hyperref}

\cvprfinalcopy 

\def\cvprPaperID{****} 

\ifcvprfinal\fi
\begin{document}

\title{A Global-Local Graph Attention Network for Traffic Forecasting}

\author{Tianchi Zhang \\
{\tt\small tonyztc@umich.edu}
}

\maketitle
\begin{abstract}
Traffic forecasting is a significant part of intelligent transportation systems. One of the critical challenges of traffic forecasting is to find spatio-temporal correlations. In recent years, graph convolutional networks and graph attention networks have replaced traditional statistical models to predict future traffic. However, it is complicated for both of them to allow vertices to have far different characters. To address this, we propose the Global-Local Graph Attention Network (GLGAT) with pairwise encoding and the event-based adjacency matrix. The GLGAT allows vertices to have a global attention matrix set for the whole graph and assigns local attention matrix sets to each vertex. Experiments on two real-world traffic datasets show that GLGAT can effectively capture spatio-temporal correlations and has competitive performance against other state-of-the-art baselines.
\end{abstract}

\section{Introduction}
Nowadays, traffic forecasting has gained more and more attention as a critical part of cars’ self-driving and route-planning systems and an essential part of urban intelligent transportation systems. With an increasing number of real-time traffic sensors on the road, more and more data become available for forecasting methods to forecast traffic congestion, find time-saving travel routes, locate bottlenecks of city traffic, and help urban planning. Moreover, with the rapid development of mobile networks and positioning systems, it has become accessible for many individuals and organizations to use real-time traffic forecasting. A forecasting algorithm with higher accuracy will be beneficial for these mentioned areas. 

The study of traffic forecasting has been going on for decades. As a prediction problem of different locations and times, one feature of the traffic forecasting problem is that models need to find the spatio-temporal correlations. Researchers used traditional statistical models, e.g., auto-regressive integrated moving average (ARIMA) \cite{Kumar_Vanajakshi_2015, WANG2005141, ZHANG2003159} and Kalman filtering, in the early stage of the study. However, as most of them use linear architectures, they are not suitable for the highly non-linear problem. 

The performance has undergone dramatic improvement as an increasing number of deep learning models have recently become practical to solve this problem \cite{jiang2021graph, QU2019304}. On the spatial correlation side, researchers treated the city as an image. They labeled sensors on their real-life locations and solved the problem with Convolutional Neural Network (CNN). Most recently, Graph Convolutional Network (GCN) and Graph Attention Network (GAT) have become mainstream since they are not limited by Euclidean spatial relationships. On the temporal correlation side, Recurrent Neural Networks (RNN) and their variants, Long Short-Term Memory (LSTM), bidirectional LSTM, and Gated Recurrent Unit (GRU), become popular sequence-to-sequence approaches in recent works. The usage of transformers, a leading trend in natural language processing (NLP), also appeared in some latest papers. Some researchers proposed methods that could identify spatio-temporal correlations simultaneously, like Graph LSTM. 

Another feature of the traffic forecasting problem is that each sensor has its character. For example, sensors on one-way roads might have more directional preferences than sensors on the cross-way, and sensors downtown might be more time-related than sensors uptown, and sensors on narrow streets might be more easily affected by surroundings than sensors on wide streets. However, recent approaches, like GCN and GAT, either fail to distinguish different adjacent sensors or demand large hidden dimensions as well as high network depth to allow each sensor to have different characters.

To address this problem, we present a new model called Global-Local Graph Attention Network (GLGAT), an extension of the GAT. More specifically, unlike regular graph attentions, which use three matrices to do self-attention, GLGAT assigns each sensor a triple of shared matrices for “global” attention and independent learnable matrices to form “local” attention functions. GLGAT supports the multi-head mechanism without losing parallelization, and it allows different adjacency matrices to act on different heads. In our experiment, GLGAT with only flatten time dimension already has a competitive performance against other state-of-the-art models with LSTM, GRU, or transformers. In conclusion, this paper has the following main contribution:
\begin{enumerate}
    \item Proposing a new graph attention framework, allowing sensors to have independent attention functions with their neighbors. It allows each sensor to have a more localized preference than the traditional graph attention network.
    \item Introducing a pairwise encoding version, which is direction-and-distance-based. Compared with other encoding methods, the new version allows sensors to have more capacity to distinguish neighborhoods by their geographic relationships.
    \item Designing a data-driven adjacency matrix based on the time correlation of speed increase and decrease events between different sensors.
\end{enumerate}

The organization of the rest of the paper is as follows. Section II is a review of related works. Section III elaborates the methodology and the structure of GLGAT. Section IV demonstrates the experiments and the results. Furthermore, we conclude in Section VI.

\section{Related Work}
In this section, we review some related literature to our work. The first half is about the graph neural networks, and the second half is about the adjacency matrices that are widely used in traffic forecasting studies.

\subsection{Graph Neural Networks}
The CNN structure has shown its promising ability to extract spatial relationships in Euclidean space in computer vision studies. Many studies adopt CNN to find the spatial correlations to beat the performance of traditional statistical models, like ARIMA and its variances \cite{Kumar_Vanajakshi_2015, WANG2005141, ZHANG2003159}. By marking regular grid maps with traffic data, spatio-temporal data becomes a series of images. Combining CNN with models that detect temporal correlations, many studies improved their prediction accuracy significantly. In our previous work \cite{Zhu_Zhang_Zhang_Zhang_2021}, the model combining CNN, LSTM, and additional meteorological data can reasonably predict taxi demand. Nevertheless, we notice that the CNN model neglects the topology structure within the sensor network, which hampers the performance when the number of sensors increases. 

The GCN is a solution that preserves both graph topological structure and convolution mechanism. Bruna et al. use Graph Laplacian in the graph convolution framework instead of the traditional square-shape convolution core. Defferrard, Bresson, and Vandergheynst use Chebyshev polynomials to reduce the computational complexity. Kipf and Welling provide a first-order Chebyshev polynomials approximation. The GCN is compatible with structures that can find the temporal correlation, like RNN, LSTM, GRU, and Transformer, using the original adjacency matrix or its higher powers. Later, noticing the achievements of attention and transformer in NLP, many researchers generalized the attention mechanism to replace the convolution core in GCN and build a new mechanism called GAT. Many studies show that replacing GCN with GAT gives performance improvements. 

The CNN, GCN, and GAT share the same idea that performing a shared transformation can extract the graph-structured information. However, in the traffic forecasting problem, sensors have their unique characters. It requires many hidden dimensions for globally shared convolution filters or attention matrices to find sensors’ local characteristics, which may cause high inter-channel redundancy. Besides, the widely used encoding, e.g., eigenvectors or sine and cosine functions, cannot represent relative relationships, which will hinder the attention mechanism from identifying sensors’ different neighbors.

\subsection{Adjacency Matrix}
Adjacency matrices are critical to numerous state-of-the-art deep learning models in traffic forecasting. A bunch of ablation studies shows that a suitable adjacency matrix can improve models’ performance. The majority of adjacency matrices are classified into three categories: connection-based, dynamic, and similarity-based. 

Connection-based matrices relate to the connectivity between sensors. The matrix values usually represent whether a road directly connects two sensors or whether vehicles can travel between them within 5 minutes. One variance of this method is to use the inverse of graph hop count or travel time to replace binary numbers, which extends the relationship to a broader neighborhood. One limitation with connection-based matrices is that they could only work with physically short-range correlations. It requires a multi-layer structure to endow models with the ability to find correlations between distant sensors. 

Unlike static matrices, dynamic matrices allow models to build the adjacency matrix while training. The initialization of the dynamic adjacency matrix is carried out through specific static methods or randomization. This method allows the model to find the most suitable matrix for the network structure. However, it requires more complex network tuning, and it makes the model less explainable. 

Similarity-based adjacency matrices have two major sub-types, specifically, functional similarity matrices and traffic pattern similarity matrices. Functional similarity matrices usually represent the point-of-interest around different sensors. This method can be helpful while forecasting the human flow of the subway, where the point-of-interest of different stations varies a lot. However, it is meaningless to predict the traffic flow of a sensor on the highway if it is in the middle of the road since cars there cannot get in or out of the road casually. Traffic pattern similarity matrices, which extracted from training data, represent the similarities of the flow patterns between sensors. For example, Dynamic Time Warping (DTW) and FastDTW \cite{10.5555/1367985.1367993} use a temporal graph to find similarities in time series. However, they require sensors to have a strong correlation during most periods. For example, it will eliminate sensor pairs with substantial similarities in a short time slot like the morning rush. 

\section{Methodology}
In this section, we formulize the traffic forecasting problem, define the graph module of traffic data, and the GAT. Then, we introduce the GLGAT as a refinement of the GAT and the pairwise encoding that complements the model. Finally, we present an event-based adjacency matrix that can fit most existing models.

\subsection{Preliminaries}
\subsubsection{Graph and Adjacency Matrix}
\noindent\indent We represent a graph $G=(V, E)$ as the topological structure of the traffic network. $V$ represents the set of sensors on the road and $\vert V \vert = N$. $E$ is the set of edges that $e\in E$ if and only if two ends of $e$ are connected directly by a road. The adjacency matrix $A\in R^{N\times N}$ presents more information about the relationship of vertices. Traditionally, $a \in A$ shows the connectivity: $a$ is zero if the row and column vertices are connected and one otherwise. In a broader definition, $A$ represents the correlations between vertices. For any $a \in A$, large $\vert a \vert$ suggests that the row vertex and the column vertex have a strong relationship, and small $\vert a \vert$ shows a weak relationship. 

\subsubsection{Traffic Forecasting Problem}
\noindent\indent The target of the traffic forecasting problem is to predict future traffic based on historical traffic data. $X^{t}\in R^{N\times K}$ denotes the $K$ traffic features observed in each sensor at time $t$. Given the graph, $G$, in $P$ previous steps of past graph signal, the traffic forecasting problem is to obtain a function $F$ that can predict the graph signal in the next $Q$ timesteps. 
\begin{equation}
	[X^{t-P+1,\dots,t}, G]\stackrel{F}{\longrightarrow}[X^{t+1,\dots,t+Q}]
\end{equation}
where $X^{t-P+1,\dots,t}\in R^{P\times N\times K}$ and $X^{t+1,\dots,t+Q}\in R^{Q\times N\times K}$. 

\subsubsection{Graph Attention Network}
\noindent\indent Veličković et al. adopted attention mechanisms to learn the coefficients between vertex pairs and proposed GAT \cite{velivckovic2017graph}. Most GAT models use the self-attention model as the base component of the graph attention layer. For graph data $X_{\mathit{in}} \in R^{N\times K}$, hidden size $H$, and output size $K’$, we can obtain the query $Q\in R^{N\times H}$, the key $K\in R^{N\times H}$, and the value $V\in R^{N\times H}$ of the self-attention as follow,
\begin{equation}
  \begin{split}
    Q &= W_{Q}(X_{\mathit{in}} \oplus E) + b_{Q}\\
    K &= W_{K}(X_{\mathit{in}} \oplus E) + b_{K}\\
    V &= W_{V}X_{\mathit{in}} + b_{V}
  \end{split}
\end{equation}
where $E \in R^{N\times H_{E}}$ is the encoding of each vertex. $\oplus$ is the concatenation operator. $W_{Q}, W_{K}\in R^{H \times (K + H_E)}$ and $W_{V}\in R^{H \times K}$ are linear transform matrices and $b_{Q}, b_{K}, b_{V}\in R^{H}$ are bias for each transformation. For two vertices $v_i$ and $v_j$, the query of $v_i$ is $q_i:= Q[i]\in R^{H}$, and the key of $v_j$ is $k_j:= K[j] \in R^{H}$. The score of $v_j$ on $v_i$ is $e_{ij}$, which can be calculated as
\begin{equation}
	e_{ij} = \texttt{GELU} (q_i \cdot k_j)
\end{equation}
where $\texttt{GELU}(\cdot)$ \cite{hendrycks2016gaussian} is the activation function introduced by Hendrycks et al., and $\cdot$ is vectors’ dot product. Then, the attention coefficient denotes as
\begin{equation}
  a_{ij} = \begin{cases}
      0, & A[i, j] = 0,\\
      \dfrac{\texttt{exp}(e_{ij})}{\sum_{v_j\in G, A[i, k]\neq 0} \texttt{exp}(e_{ik})}, & \text{otherwise},
    \end{cases}
\end{equation}
which is constrained by the adjacency matrix $A$. Then, in the hidden graph data $X_{\mathit{hidden}} \in R^{N\times H}$, the feature of $v_i$, $x’_i := X_{\mathit{hidden}} [i]$, can be computed by weighted summation of values in $V$:
\begin{equation}
	x’_i = \sum_{v_j\in G} a_{ij} V[j].
\end{equation}
Then apply a feed-forward layer to the $X_{\mathit{hidden}}$ with a transform matrix $W_{\mathit{ff}} \in R^{K’ \times H}$ and a bias $b_{\mathit{ff}} \in R^{K’}$
\begin{equation}
  X_{\mathit{out}} = W_{\mathit{ff}} X_{\mathit{hidden}} + b_{\mathit{ff}}
\end{equation}
Since $X_{\mathit{in}}$ and $X_{\mathit{out}}$ have the same graph structure, stacking GAT can create a deep neural network as long as hidden dimensions match. 

\subsection{Global-Local Graph Attention Network}
The GLGAT has a similar structure as GAT, which first transforms the input into the hidden state by a layer using the attention mechanism and then uses feed-forward layers to compute the output by the hidden state. Keeping the same feed-forward structure, the significant change of GLGAT is in the attention layer.

To support the multi-head mechanism and further allow different adjacency matrices to act on different heads, we increase the hidden size to $H’$ that
\begin{equation}
	H’ := H \times H_{\mathit{adj}} \times H_{\mathit{head}} 
\end{equation}
where $H$ is the hidden size of each head, $H_{\mathit{adj}}$ is the number of adjacency matrices used, and $H_{\mathit{head}}$ is the number of heads corresponding to each adjacency matrix. 

To allow each vertex to have its own local attention operation, the easiest way is to have $N$ independent attention operating on all vertices. However, it will increase the computational complexity about $N$ times. As a trade-off, we keep the generation procedure of key $K$ and value $V$ unchanged and assign a transforming matrix of query for each vertex. Thus the output size of query $Q$ is changed to
\begin{equation}
 H_Q := H’ + H_{\mathit{adj}} \cdot H_{\mathit{PE}},
\end{equation}
where $H_{\mathit{PE}}$ is the size of pairwise encoding $\mathit{PE} \in R^{N\times N\times H_{\mathit{PE}}}$ which will be discussed in the next sub-section. We also assign independent matrices and biases to each vertex. More precisely, for the vertex $v_i$, the query $q_i := Q[i]$ corresponding to it is generated by two parts, $q_{i, \mathit{Global}} \in R^{H_Q}$, which is calculated by “globally” shared parameters, and $q_{i, \mathit{Local}} \in R^{H_Q}$, which is calculated by “locally” owned parameters. The formulas are
\begin{equation}
  \begin{split}
    q_{i, \mathit{Global}} &= W_{\mathit{Q-Global}}(x_i \oplus e_i) + b_{\mathit{Q-Global}}\\
    q_{i, \mathit{Local}} &= W_{\mathit{Q-Local}, i}(x_i \oplus e_i) + b_{\mathit{Q-Local}, i}
  \end{split}
\end{equation}
where the $e_i := E[i]$ is the $i$-th row of the traditional encoding $E\in R^{N\times H_E}$ that matches the vertex $v_i$. The $W_{\mathit{Q-Global}} \in R^{H_Q \times (K + H_E)}$ and $b_{\mathit{Q-Global}} \in R^{H_Q}$ are shared transform matrices and biases. The $W_{\mathit{Q-Local}, i} := W_{\mathit{Q-Local}}[i]$ and $b_{\mathit{Q-Local},i} := b_{\mathit{Q-Local}}[i]$ are $v_i$’s private transforming matrices and biases to extract unique features for each vertex, and $W_{\mathit{Q-Local}} \in R^{N \times H_Q \times (K + H_E)}$, $b_{\mathit{Q-Local}} \in R^{N \times H_Q}$. The query $q_i$ is determined by compressing $q_{i, \mathit{Global}}$ and $q_{i, \mathit{Local}}$ as
\begin{equation}
  q_i = W_Q (q_{i, Global} \oplus q_{i, Local}) + b_Q
\end{equation}
where $W_Q \in R^{H_Q \times (2H_Q)}$, $b_Q \in R^{H_Q}$ are transforming parameters. The query $Q \in R^{N \times H_Q}$ can be obtained by stacking $q_i$s.  By splitting the last dimension, the query will be separated to two parts, the query for attention $Q_{\mathit{AT}} \in R^{N \times H’}$ and the query for pairwise encoding $Q_{\mathit{PE}} \in R^{N \times (H_{\mathit{adj}} \times H_{\mathit{PE}})}$.

Then reshape all matrices to fit the multi-head structure, i.e., splitting the last dimension of the graph data. The detailed shape changes are
\begin{equation}
  \begin{split}
    Q_{\mathit{AT}}:\quad & (N \times H’) \longrightarrow (N \times H_{\mathit{adj}} \times H_{\mathit{head}} \times H),\\
    Q_{\mathit{PE}}:\quad & (N \times (H_{\mathit{adj}} \cdot H_{\mathit{PE}})) \longrightarrow (N \times H_{\mathit{adj}} \times H_{\mathit{PE}}),\\
    K:\quad & (N \times H’) \longrightarrow (N \times H_{\mathit{adj}} \times H_{\mathit{head}} \times H),\\
    V:\quad & (N \times H’) \longrightarrow (N \times H_{\mathit{adj}} \times H_{\mathit{head}} \times H).
  \end{split}
\end{equation}
For the $n$-th adjacency matrix $A_n$, for its $m$-th head, the score of $v_j$ on $v_i$ can be calculated by
\begin{equation}
  e_{ij, n, m} = \texttt{GELU} (q_{\mathit{at}, i, n, m} \cdot k_{j, n, m} + q_{\mathit{pe}, i, n} \cdot \mathit{pe}_{i, j}),
\end{equation}
where $q_{\mathit{at}, i, n, m} := Q_{\mathit{AT}}[i, n, m] \in R^{H}$, $k_{j, n, m} := K[j, n, m] \in R^{H}$, $q_{\mathit{pe}, i, n} := Q_{\mathit{PE}}[i, n] \in R^{H_{\mathit{PE}}}$, and $\mathit{pe}_{i, j} := \mathit{PE}[i, j] \in R^{H_{\mathit{PE}}}$. The attention coefficient denotes as
\begin{equation}
  a_{ij, n, m} = \dfrac{ \texttt{exp} (e_{ij, n, m}) \cdot A_{n}[i, j]}{\sum_{v_j\in G} \texttt{exp}(e_{ik, n, m}) \cdot A_{n}[i, k]},
\end{equation}
where $A_{n}[i, j] \in [0, 1]$ is the value in $i$-th row and $j$-th column of the $n$-th adjacency matrix. Using the attention coefficient as weights and sum up the values, we have the hidden graph data $X_{\mathit{hidden}} \in R^{N \times H_{\mathit{adj}} \times H_{\mathit{head}} \times H}$ obtained by
\begin{equation}
	x_{\mathit{hidden}, i, n, m} = \sum_{v_j\in G} a_{ij, n, m} v{j, n, m}.
\end{equation}
where $v{j, n, m} := V[j, n, m]$. 

By flattening the last three dimensions, we have
\begin{equation}
  X_{\mathit{hidden}}: (N \times H_{\mathit{adj}} \times H_{\mathit{head}} \times H) \longrightarrow (N \times H’)
\end{equation}
Then, using the transform matrix $W_{\mathit{ff}} \in R^{K’ \times H’}$ and bias $b_{\mathit{ff}} \in R^{K’}$, the feed-forward layer will generate $X_{\mathit{output}}$ same as the traditional way. As long as the dimensions match, stacking GLGAT will produce a deeper neural network since the graph structure is preserved during the transformation. 

\subsection{Pairwise Encoding}
Introducing pairwise encoding $\mathit{PE} \in R^{N \times N \times H_{\mathit{PE}}}$ is to provide more information during the query-key comparison step. The previous subsection shows that the $\mathit{PE}$ acts like a key supplement and operates with a separate query. The pairwise encoding stores some basic correlations between vertices. A $H_{\mathit{PE}}$ dimension data of the comparison of $v_i$ and $v_j$ can be stored in $\mathit{PE}[i, j]$. 

Several methods can form the encoding. Here we demonstrate a direction-and-distance-based version of pairwise encoding. The traffic flow on one sensor is usually influenced by two factors, one is possible incidents along the road where the sensor is located, another one is the chosen paths and destinations of people passing this sensor. Both of them show that the strength of correlation between sensors is subject to the direction of the road. A function $\mathit{ED}(\cdot)$ transform position to the direction and classify all directions into eight classes, forty-five degrees each, and encode with one-hot-coding with label-smoothing. For two vertices $v_i$ and $v_j$, whose corresponding sensors’ locations are $(x_i, y_i)$ and $(x_j, y_j)$, the $\mathit{pe}_{ij} := \mathit{PE}[i, j]$ and $\mathit{pe}_{ji} := PE[j, i]$ are formulized by $\mathit{ED}(\cdot)$ and distance functions:
\begin{align}
  \begin{split}
    &E_{\mathit{direction}, ij} = \mathit{ED}(x_i, y_i, x_j, y_j),\\
    &E_{\mathit{direction}, ij} = \mathit{ED}(x_i, y_i, x_j, y_j),\\
    &E_{L1} = \vert x_i - x_j \vert + \vert y_i - y_j \vert, \\
    &E_{L2} = \sqrt{(x_i - x_j)^2 + (y_i - y_j)^2}.
  \end{split}\\
  \mathit{pe}_{ij} &= E_{\mathit{direction}, ij} \oplus E_{L1} \oplus E_{L2},\\
  \mathit{pe}_{ji} &= E_{\mathit{direction}, ji} \oplus E_{L1} \oplus E_{L2}.
\end{align}

Other information like the point-of-interest difference of each sensor’s surroundings can also be encoded in the pairwise encoding. The pairwise encoding can also show sensors’ relative sequential information by comparing distances between sensors and highway exits in each sensor pair. Besides, traffic forecasting problems in different timesteps can have different $\mathit{PE}$s, where the historical traffic flow between each sensor during the same time of the day can be embedded. A dynamic pairwise encoding is also allowed like traditional encoding.

\subsection{Event-Based Adjacency Matrix}
The event-based adjacency matrix method is a data-driven approach designed for the traffic forecasting problem. The main idea is that, for one certain sensor, we find the other sensors that are likely to have a similar traffic flow change which would happen just a few time-steps before the chosen sensor. First, define a “divider” and set its value as the average value of the maximum and minimum historical traffic flow for each sensor. Then, define the “up-event,” which only happens if the traffic flow increases and surpasses the “divider” value. Similarly, define the “down-event” only if the traffic flow decreases and passes through the “divider” value. The set of events of vertex $v_i$ that corresponding to the sensor is called $\mathit{EV}_{i, \mathit{up}}$ and $\mathit{EV}_{i, \mathit{down}}$ correspondingly,
\begin{equation}
  \begin{split}
    \mathit{EV}_{i, \mathit{up}} &= [t_{\mathit{up}, 1}, \dots, t_{\mathit{up}, u}],\\
    \mathit{EV}_{i, \mathit{down}} &= [t_{\mathit{down}, 1}, \dots, t_{\mathit{down}, d}].
  \end{split}
\end{equation}

For each $t_{\mathit{up}, n} \in \mathit{EV}_{i, \mathit{up}}$, we find all $v_j$ have “up-event” that is contiguous to $t_{up, n}$, such that 
\begin{equation}
  [t_{\mathit{up}, n} - t_p, \dots, t_{\mathit{up}, n} + t_q] \cap \mathit{EV}_{j, \mathit{up}} \neq \emptyset,
\end{equation}
and group them as $\mathit{Group}(t_{\mathit{up}, n})$, where $t_p$ and $t_q$ determine acceptable range. Then all related vertex group $\mathit{VG}$ of $v_i$ is
\begin{equation}
  \mathit{VG}_{\mathit{up}, i} = [\mathit{Group}(t_{\mathit{up}, i}), \dots, \mathit{Group}(t_{\mathit{up}, u})].
\end{equation}
Similarly, we have
\begin{equation}
  \mathit{VG}_{\mathit{down}, i} = [\mathit{Group}(t_{\mathit{down}, i}), \dots, \mathit{Group}(t_{\mathit{down}, u})].
\end{equation}
Based on the frequency of each vertex appearing in $\mathit{VG}_{\mathit{up}, i}$ and $\mathit{VG}_{\mathit{down}, i}$, the $i$-th row of a new adjacency matrix can be determined. Both the adjacency matrix of “up-event” and adjacency matrix of “down-event” are included in our model.

\section{Experiments}
In this section, we first introduce the datasets and baselines that we use. Then the setup of our model is shown, followed by the result of the experiments. Finally, we conduct a detailed ablation study.
\subsection{Datasets}
Our experiments are performed on two typical real-world datasets, METR-LA and PEMS-BAY \cite{li2017diffusion}.
\begin{enumerate}
\item METR-LA: This dataset is collected from loop detectors in the Los Angeles County road network. It contains 207 sensors’ data in four months from March 1st, 2012 to June 30th, 2012, i.e., 34,272 5-minute slices included.
\item PEMS-BAY: This dataset is collected by California Transportation Agencies Performance Measurement System. It contains 325 sensors’ data in six months from January 1st, 2017 to May 31st, 2017, i.e., 52,116 5-minute slices included.
\end{enumerate}
Some key features of two datasets are summarized in the table \ref{tab:dataset}.
\begin{table}
\begin{center}
\begin{tabular}{lcccc}
\hline
\hline
Dataset &  Sensors & Size & Unit & Missing Ratio \\
\hline
METR-LA   & 207 & 34272 & 5 min & 8.109\%\\
PEMS-BAY & 325  & 52116 & 5 min & 0.003\% \\
\hline
\hline
\end{tabular}
\caption{Dataset Statistics.}
\label{tab:dataset}
\end{center}
\end{table}

\subsection{Baselines}
We compare the GLGAT with other traffic forecasting models, including
\begin{enumerate}
\item HA: Historical Average model uses the weighted average of historical data of the same time-of-the-day to predict the future traffic flow.
\item ARIMA: Auto-regressive Integrated Moving Average model \cite{Kumar_Vanajakshi_2015, WANG2005141, ZHANG2003159} with Kalman filter is a classic time series analysis model using a linear architecture.
\item VAR: Vector Auto-Regression \cite{6482260, lutkepohl2005new} evaluates the relationships between time series, assuming that they are stationary.
\item SVR: Support Vector Regression \cite{Smola_2004} adapts the support vector machine model for traffic data.
\item FNN: Feed Forward Neural Network directly flattens the input data and deals with traffic sequences.
\item FC-LSTM: Fully Connected Long Short Term Memory network \cite{sutskever2014sequence} uses LSTM structure to handle data from different timesteps.
\item DCRNN: Diffusion Convolutional Recurrent Neural Network \cite{li2017diffusion} uses bidirectional random works for spatial correlation and encoder-decoder architecture for temporal dependency. 
\item STGCN: Spatial-Temporal Graph Convolutional Network \cite{Yu_Yin_Zhu_2018} is a combination of graph convolutional and one-dimension convolution.
\item GaAN: Gated Attention Networks \cite{zhang2018gaan} uses an attention-based network with a convolutional sub-network and gate mechanism.
\item APTN: Attention-based Periodic-Temporal neural Network \cite{9062547} uses the encoder attention mechanism to catch spatial and periodical dependencies.
\item GST-GAT: Global Spatial–Temporal Graph Attention Network uses \cite{9316302} the graph attention with LSTM and gating fusion mechanism.
\end{enumerate}
The first four methods, HA, ARIMA, VAR, and SVR, are traditional statistical models. The FNN, FC-LSTM, DCRNN, and STGCN are deep-learning models that do not use attention mechanisms. The last three models, GaAN, APTN, and GST-GAT, are attention-based.

\subsection{Experiment Setup}
The experiment setup follows the setting in the paper of DCRNN \cite{li2017diffusion}. 70\% of the data is the training set, 10\% are the cross-validation set, and the remaining 20\% are the testing set. Each sample sequence contains two-hour data, with 5-minute intervals. Thus there are twenty-four timesteps divided into two halves. $P$ and $Q$ defined in the previous section are all twelve, i.e., the first 12 timesteps are the input, and the rest 12 timesteps are the ground truth that models need to predict. 

The GLGAT model testing has a seven-layer structure. A network block is shared on each floor. The bottom layer uses GLGAT blocks of input size as three timesteps and output size 16. It pads the data in the last timestep twice, and every three adjacent timesteps are grouped. The second layer’s GLGAT blocks have input and output sizes equal to 16. The third layer is a time-flatten layer where the data of different timesteps are concatenated into a vector of length 192. The fourth, fifth, and sixth layers are GLGAT blocks of input size and output size equal to 192. Then the last layer is a fully connected layer which compresses the data dimension to 12 and outputs them as predictions of twelve future timesteps. The initial learning rate is $1.0\times 10^{-4}$. We optimize the model with Adam optimization by minimizing the smooth L1 loss \cite{girshick2015fast} between $Y_i$ and $\hat{Y}_i$. Our experiments run on a Windows computer with one Intel(R) Core(TM) i7-10750H CPU, 32GB RAM, and one NVIDIA GeForce RTX 2080 Super GPU. 

Three conventional matrices, Mean Absolute Error (MAE), Root Mean Square Error (RMSE), and Mean Absolute Percentage Error (MAPE), measure the performance of different models. Their formulas are
\begin{align}
    \mathit{MAE} &= \dfrac{1}{N}\sum^{N}_{i=1} \left\vert Y_i - \hat{Y}_i \right\vert, \\
    \mathit{RMSE} &= \sqrt{  \dfrac{1}{N}\sum^{N}_{i=1} \left( Y_i - \hat{Y}_i \right)^2}, \\
  \mathit{MAPE} &= \dfrac{100\%}{N}\sum^{N}_{i=1} \left\vert \dfrac{Y_i - \hat{Y}_i}{Y_i} \right\vert,
\end{align}
where $N$ is the number of testing, $Y_i$ is the ground truth of the $i$-th sample, and $\hat{Y}_i$ is the prediction from the model. Smaller MAE, MAPE, and RMSE values suggest better performance.

\subsection{Experiment Results}
\begin{table*}
\begin{center}
\begin{tabular}{l|ccc|ccc|ccc}
\hline
\hline
\multirow{2}*{Models} & \multicolumn{3}{|c|}{15 minutes} & \multicolumn{3}{|c|}{30 minutes} & \multicolumn{3}{|c}{60 minutes} \\
\cline{2-10}
& MAE & RMSE & MAPE & MAE & RMSE & MAPE & MAE & RMSE & MAPE \\
\hline
HA &  4.16 & 7.80 & 13.0\% & 4.16 & 7.80 & 13.0\% & 4.16 & 7.80 & 13.0\% \\
ARIMA & 3.99 & 8.21 & 9.6\% & 5.15 & 10.45 & 12.7\% & 6.90 & 13.23 & 17.4\% \\
VAR & 4.42 & 7.89 & 10.2\% & 5.41 & 9.13 & 12.7\% & 6.52 & 10.11 & 15.8\% \\
SVR & 3.99 & 8.45 & 9.3\% & 5.06 & 10.87 & 12.1\% & 6.72 & 13.76 & 16.7\% \\
\hline
FNN & 3.99 & 7.94 & 9.9\% & 4.23 & 8.17 & 12.9\% & 4.49 & 8.69 & 14.0\% \\
FC-LSTM & 3.44 & 6.30 & 9.6\% & 3.77 & 7.23 & 10.9\% & 4.37 & 8.96 & 13.2\% \\
DCRNN & 2.77 & 5.38 & 7.3\% & 3.15 & 6.45 & 8.8\% & \underline{3.60} & \underline{7.59} & \underline{10.5\%} \\
STGCN & 2.88 & 5.74 & 7.62\% & 3.47 & 7.24 & 9.57\% & 4.59 & 9.40 & 12.70\% \\
\hline
GaAN & \underline{2.71} & 5.24 & \underline{6.99\%} & 3.12 & \underline{6.36} & \underline{8.56\%} & 3.64 & 7.65 & 10.62\% \\
APTN & 2.76 & 5.38 & 7.30\% & 3.15 & 6.43 & 8.80\% & 3.70 & 7.69 & 10.69\% \\
GST-GAT & \textbf{2.51} & \textbf{5.23} & 7.25\% & \textbf{3.07} & 6.55 & 8.57\% & 4.10 & 8.16 & 11.89\% \\
\hline
GLGAT & 2.73 & \textbf{5.23} & \textbf{6.95\%} & \underline{3.08}  & \textbf{6.18}  &  \textbf{8.33\%} & \textbf{3.52}  & \textbf{7.31}  &  \textbf{10.23} \% \\
\hline
\hline
\end{tabular}
\captionsetup{width=.85\textwidth}
\caption{Traffic Prediction on the METR-LA. Bold numbers suggest best performance among these models, and underlined numbers means the second-best.}
\label{tab:metr-la-result}
\end{center}
\end{table*}

\begin{table*}
\begin{center}
\begin{tabular}{l|ccc|ccc|ccc}
\hline
\hline
\multirow{2}*{Models} & \multicolumn{3}{|c|}{15 minutes} & \multicolumn{3}{|c|}{30 minutes} & \multicolumn{3}{|c}{60 minutes} \\
\cline{2-10}
& MAE & RMSE & MAPE & MAE & RMSE & MAPE & MAE & RMSE & MAPE \\
\hline
HA & 2.88 & 5.59 & 6.8\% & 2.88 & 5.59 & 6.8\% & 2.88 & 5.59 & 6.8\% \\
ARIMA & 1.62 & 3.3 & 3.5\% & 2.33 & 4.76 & 5.4\% & 3.38 & 6.5 & 8.3\% \\
VAR & 1.74 & 3.16 & 3.6\% & 2.32 & 4.25 & 5.0\% & 2.93 & 5.44 & 6.5\% \\
SVR & 1.85 & 3.59 & 3.8\% & 2.48 & 5.1 & 5.5\% & 3.28 & 7.08 & 8.0\% \\
\hline
FNN & 2.2 & 4.42 & 5.19\% & 2.3 & 4.63 & 5.43\% & 2.46 & 4.98 & 5.89\% \\
FC-LSTM & 2.05 & 4.19 & 4.8\% & 2.20 & 4.55 & 5.2\% & 2.37 & 4.96 & 5.7\% \\
DCRNN & \underline{1.31} & \underline{2.76} & \underline{2.74\%} & 1.66 & 3.78 & 3.76\% & 1.98 & 4.62 & 4.74\% \\
STGCN & 1.36 & 2.96 & 2.90\% & 1.81 & 4.27 & 4.17\% & 2.49 & 5.69 & 5.79\% \\
\hline
GaAN & - & - & - & - & - & - & - & - & - \\
APTN & 1.38	& 2.96 & 2.91\% & 1.97 & 3.95 & 3.69\% & 2.33 & 4.60 & 4.65\% \\
GST-GAT & \textbf{1.28} & \textbf{2.65} & \textbf{2.64\%} & \textbf{1.48} & \textbf{3.52} & \textbf{3.41\%} & \textbf{1.92} & \underline{4.41} & \textbf{4.39\%} \\
\hline
GLGAT & \underline{1.31} & \underline{2.76} & 2.75\% & \underline{1.63} & \underline{3.67} & \underline{3.62\%} & \textbf{1.92} & \textbf{4.36} & \underline{4.50\%} \\
\hline
\hline
\end{tabular}
\captionsetup{width=.85\textwidth}
\caption{Traffic Prediction on the PEMS-BAY. Bold numbers suggest best performance among these models, and underlined numbers means the second-best.}
\label{tab:pems-bay-result}
\end{center}
\end{table*}

The Table \ref{tab:metr-la-result} and Table \ref{tab:pems-bay-result} shows the performance matrices of each model on 15 minutes, 30 minutes, and 60 minutes ahead prediction, where traditional statistical models, deep-learning models without attention, and deep-learning models with attention are separated by horizontal lines. 

HA uses traffic flow patterns in the training set and does not consider the immediate past data when testing. Unlike other models, HA’s performance is irrelevant to the increase in the forecasting horizon. It has relatively good performance for long-term prediction, but it is not capable of handling sudden changes. Other traditional statistical methods have better short-term prediction accuracy but worse 60-minute forecasting results than HA.

The deep-learning models with and without attention mechanisms have better performance than traditional statistical models. Moreover, attention-based models have comparable performance compared with other deep-learning models. In the METR-LA dataset, our model, GLGAT, listed at the bottom of two tables, achieves improvement over other methods in a majority number of matrices. In the PEMS-BAY, most of the results of GLGAT are the best or the second-best from all comparing methods.

\subsection{Ablation Study}
Additional experiments are designed to verify the effect of each component. We design three ablation models:
\begin{enumerate}
  \item Ablation-1: Replace the event-based adjacency matrices with the adjacency matrices of connectivity.
  \item Ablation-2: Remove the pairwise encoding.
  \item Ablation-3: Replace the GLGAT with traditional GAT and remove the pairwise encoding.
\end{enumerate}

\begin{table*}
\begin{center}
\begin{tabular}{l|ccc|ccc|ccc}
\hline
\hline
\multirow{2}*{Models} & \multicolumn{3}{|c|}{15 minutes} & \multicolumn{3}{|c|}{15 minutes} & \multicolumn{3}{|c}{60 minutes} \\
\cline{2-10}
& MAE & RMSE & MAPE & MAE & RMSE & MAPE & MAE & RMSE & MAPE \\
\hline
GLGAT & \textbf{2.73} & \textbf{5.23} & \textbf{6.95\%} & \textbf{3.08}  & \textbf{6.18}  &  \textbf{8.33\%} & \textbf{3.52}  & \textbf{7.31}  &  \textbf{10.23} \% \\
\hline
Ablation-1 & 2.74  & 5.28  & 7.18\% & 3.13  &  6.41 &   8.79\% &  3.64 &  7.61 &   10.38\% \\
Ablation-2 & 2.76  & 5.31  &   7.19\% & 3.16  & 6.44  &   8.69\% & 3.65  &  7.64 &   10.45\% \\
Ablation-3 &  2.80 &  5.47 &  7.23\% & 3.22  &  6.65 &  8.83\%  & 3.76  & 7.91  &  10.85\%\\
\hline
\hline
\end{tabular}
\captionsetup{width=.85\textwidth}
\caption{Ablation study on the METR-LA. Bold numbers suggest best performance among these models.}
\label{tab:metr-la-ablation}
\end{center}
\end{table*}

\begin{table*}
\begin{center}
\begin{tabular}{l|ccc|ccc|ccc}
\hline
\hline
\multirow{2}*{Models} & \multicolumn{3}{|c|}{15 minutes} & \multicolumn{3}{|c|}{15 minutes} & \multicolumn{3}{|c}{60 minutes} \\
\cline{2-10}
& MAE & RMSE & MAPE & MAE & RMSE & MAPE & MAE & RMSE & MAPE \\
\hline
GLGAT & \textbf{1.31} & \textbf{2.76} & \textbf{2.75\%} & \textbf{1.63} & \textbf{3.67} & \textbf{3.62\%} & \textbf{1.92} & \textbf{4.36} & \textbf{4.50\%} \\
\hline
Ablation-1 &  1.34 &  2.85 &  2.90\% &  1.65 &  3.74 &   3.76\% & 1.97  &  4.53 &  4.55 \% \\
Ablation-2 & 1.36  &  2.82 &  2.88\% &  1.68 &  3.77 &  3.80\% & 1.96  & 4.46 &   4.68\% \\
Ablation-3 & 1.38  & 2.89  & 2.87\% & 1.69  &  3.70 &   3.78\% &  2.08 &  4.60 &   4.78\% \\
\hline
\hline
\end{tabular}
\captionsetup{width=.85\textwidth}
\caption{Ablation study on the PEMS-BAY. Bold numbers suggest best performance among these models.}
\label{tab:pems-bay-ablation}
\end{center}
\end{table*}

The performance of GLGAT and four ablation studies is in the Table \ref{tab:metr-la-ablation} and Table \ref{tab:pems-bay-ablation}. The difference between GLGAT and Ablation-1 shows the efficiency of the event-based adjacency matrices. The improvement from Ablation-3 to Ablation-2 demonstrates the potency of the GLGAT structure. Furthermore, the improvement from Ablation-2 to the full GLGAT suggests that the pairwise encoding fits the network structure. 

\section{Conclusion}
This paper proposes GLGAT, a deep learning framework for the traffic forecasting problem. Along with a set of globally shared attention matrices, GLGAT allows each vertex to have specialized local attention matrices. The pairwise encoding and the event-based adjacency matrix come up as supplements of GLGAT structure. Experiments show that GLGAT has competitive performance on two real-world datasets against other state-of-the-art models. It is worth noting that this performance is achieved without using state-of-the-art structures of finding temporal correlations, like LSTM, GRU, or transformer. For future works, we will try to combine GLGAT with other models and reinforce the model with additional meteorological data.

{\small
\bibliographystyle{ieee_fullname}
\bibliography{egbib}
}

\end{document}